\documentclass[lettersize,journal]{IEEEtran}

\usepackage{graphicx}
\usepackage{amsmath}
\usepackage{amssymb}
\usepackage{amsthm}
\usepackage[ruled,linesnumbered]{algorithm2e}
\usepackage{multirow}
\usepackage{xcolor}
\usepackage{float}
\usepackage{gensymb}
\usepackage{hyperref}
\usepackage{dblfloatfix}
\usepackage{soul}
\usepackage{pdflscape}
\usepackage{import}
\usepackage{booktabs}
\usepackage{tikz}
\usetikzlibrary{shapes.geometric, shapes, arrows, positioning, fit, calc}
\usepackage{pdflscape}
\usepackage{ragged2e}
\usepackage{framed}
\usepackage{multicol}
\usepackage{nomencl}
\usepackage{etoolbox}
\usepackage{changepage} 
\usepackage{bm}
\usepackage{orcidlink} 
\hypersetup{
    colorlinks=true,
    urlcolor=blue,
    citecolor=red
} 

\begin{document}

\title{\centering IRL-DAL: Safe and Adaptive Trajectory Planning for Autonomous Driving via Energy-Guided Diffusion Models} 
\vspace{0.2in}
\author{
    Seyed Ahmad Hosseini Miangoleh\textsuperscript{\scalebox{1.25}{\orcidlink{0009-0007-6572-6688}} 1},  
    Amin Jalal Aghdasian\textsuperscript{\scalebox{1.25}{\orcidlink{0009-0003-8482-1219}} 1}, 
    Farzaneh Abdollahi\textsuperscript{\scalebox{1.25}{\orcidlink{0000-0003-4957-987X}} 1}\\
    \textsuperscript{1}Department of Electrical Engineering, Amirkabir University of Technology (Tehran Polytechnic), Tehran, Iran
}

\maketitle

\begin{abstract}
This paper proposes a novel inverse reinforcement learning framework using a diffusion-based adaptive lookahead planner (IRL-DAL) for autonomous vehicles. Training begins with imitation from an expert finite state machine (FSM) controller to provide a stable initialization. Environment terms are combined with an IRL discriminator signal to align with expert goals. Reinforcement learning (RL) is then performed with a hybrid reward that combines diffuse environmental feedback and targeted IRL rewards. A conditional diffusion model, which acts as a safety supervisor, plans safe paths. It stays in its lane, avoids obstacles, and moves smoothly. Then, a learnable adaptive mask (LAM) improves perception. It shifts visual attention based on vehicle speed and nearby hazards. After FSM-based imitation, the policy is fine-tuned with Proximal Policy Optimization (PPO). Training is run in the Webots simulator with a two-stage curriculum. A 96\% success rate is reached, and collisions are reduced to 0.05 per 1k steps, marking a new benchmark for safe navigation. By applying the proposed approach, the agent not only drives in lane but also handles unsafe conditions at an expert level, increasing robustness.We make our code publicly available\footnote{\href{https://seyed07.github.io/Autonomous-Driving-via-Hybrid-Learning-and-Diffusion-Planning/}{Paper Website}}

\end{abstract}


\section{Introduction}
The main challenge in autonomous vehicles is building systems that can operate at human levels of safety and reliability in highly dynamic environments \cite{crosato2024social,kiran2021deep}. This challenge mainly comes from the need to avoid obstacles in a strong and reliable way. Obstacle avoidance is the key safety task that shows whether a system can work in the real world. Even small mistakes in rare situations can cause very serious failures. This means the agent must remain safe even when encountering situations it was not trained on \cite{ross2011reduction}. To address these challenges, recent research has investigated a range of advanced methods. These approaches aim to enhance the safety, reliability, and adaptability of autonomous systems operating in dynamic environments.

\subsection{Related Work}
\label{sec:related_work}
This work falls within the overlap of four main areas: hybrid learning, reward inference, generative planning, and adaptive perception. The following section reviews some recent studies in each of these areas.

\subsubsection{Hybrid Imitation and Reinforcement Learning}
Imitation Learning (IL), especially Behavioral Cloning (BC), is widely used in autonomous driving \cite{cui2023safe}. It provides a data-efficient way to learn a mapping from expert demonstrations to control actions using supervised learning \cite{li2022driver,wang2024implicit}. The primary advantage of BC is its computational efficiency and its requirement for no explicit knowledge of the underlying environment dynamics. Within BC, methods such as Conditional Imitation Learning improve performance by conditioning the policy on high-level commands. This helps address the challenge of having many possible correct actions for a given situation \cite{eraqi2022dynamic}. Even though it is efficient, IL has a problem called covariate shift. In this case, small mistakes grow over time when the agent reaches states that were not in the training data. Unlike traditional methods, RL can develop strong and resilient behaviors by learning from trial-and-error experience. However, it usually needs a lot of data and depends on hand-designed reward signals that can be unreliable \cite{yuan2025model}. 

The hybrid method combines IL and RL. This lets the self-driving car first learn basic behavior from examples and then improve it through trial, feedback, and optimization \cite{radwan2021obstacles}. In \cite{mahmoudi2023reinforcement}, a hybrid method is used that combines BC with the PPO algorithm in the Unity Agents environment. This setup trains racecar agents to drive and avoid obstacles. These combinations capitalize on expert supervision for fast convergence while allowing the agent to explore recovery and adaptation strategies \cite{pinto2021curriculum}. In \cite{lu2023imitation}, the BC-SAC model fuses BC and Soft Actor-Critic to enhance policy robustness and safety in autonomous driving. By employing supervised imitation and reinforcement optimization, one achieves higher generalization and a 38\% reduction in failure rate across complex real-world scenarios.

\subsubsection{Inverse Reinforcement Learning for Reward Shaping}
The problem of designing a good reward for complex tasks can be handled with Inverse Reinforcement Learning (IRL). IRL tries to find the hidden reward rules directly from expert demonstrations \cite{zhao2024survey,lanzaro2025evaluating}. The Conditional Predictive Behavior Planning model combines Conditional Motion Prediction and Maximum Entropy IRL to make driving more like a human. It predicts how nearby cars will react to each possible move and scores these moves using expert driving data \cite{huang2023conditional}. Adversarial inverse reinforcement learning (AIRL) uses a GAN to learn the reward and the driving policy at the same time, which helps the agent adapt to new environments. Safety-aware AIRL then adds safety rules to block risky actions and lower the chance of crashes \cite{wang2021decision}.

\subsubsection{Adaptive Perception and Attention in Driving}
With the growing use of attention mechanisms in deep learning \cite{lu2024epitester,xi2023ema}, these methods have been added to end-to-end driving models. They help the system focus more on key visual elements such as vehicles, pedestrians, and road signs \cite{wang2024pedestrian}. Effective driving requires perception systems that can adapt attention dynamically to changing contexts. In \cite{schwonberg2023survey}, it uses spatiotemporal, uncertainty-aware attention over multimodal sensors with crossmodal alignment and multiscale fusion to prioritize important actors and regions for downstream planning. In \cite{chi2024dynamic}, it adds a temporal residual block, multiscale feature fusion, and global plus double attention to use time and image cues better. In \cite{ma2024adaptive}, it combines adaptive channel attention and grouped spatial attention with channel shuffle to highlight important features. It plugs into standard CNNs, can replace a 3×3 convolution layer, and improves accuracy with little extra compute.

\subsubsection{Diffusion Models for Safe Motion Planning}
Diffusion models \cite{peng2025diffusion} are now leading tools for generating data. They are also being used more and more for planning tasks. When generating trajectories, they can produce diverse future motions that still obey real-world physics \cite{wang2024trajectory}. Their key advantage is their flexibility. By guiding the backward diffusion steps with extra rules or helper models, they can add safety rules such as avoiding crashes and keeping motions that the vehicle can really follow \cite{pearce2023plannable_journal, chi2023diffusion}. Current methods use diffusion models as direct policy models, mainly in offline reinforcement learning settings \cite{wang2023diffusion}. Moreover, some methods use stand-alone planners that first produce open-loop trajectories. A separate controller then follows these trajectories \cite{janner2022planning}.

\subsection{Research Gap}
\label{sec:gap}

Despite progress in each area, a key gap appears at their intersection. Current stacks rarely combine generative planning, reward inference, and online policy learning into one system. This causes three main limits:

\begin{enumerate}
	\item Lack of an end-to-end unified loop: Most diffusion-based planners generate trajectories open loop and are executed separately from the RL policy, creating a distribution mismatch between planned motions and closed-loop control. This separation weakens robustness under disturbances and hinders consistent transfer from trajectory proposals to joint steering–speed commands.
	\item Non-adaptive safety trade-offs: Fixed cost weights for lane keeping, collision avoidance, and stability cannot re-balance as scene risk and sensor uncertainty change. Without real-time, perception-conditioned guidance, systems oscillate between over-conservative and overly aggressive behavior and fail to maintain safety while preserving efficiency.
	\item Inefficient learning signals: Reliance on sparse, hand-crafted rewards leads to sample-hungry training, unstable convergence, and sub-expert driving. The absence of dense rewards inferred from demonstrations and a staged curriculum limits generalization to out-of-distribution states and degrades control smoothness.
\end{enumerate}

\subsection{Motivation}
\label{sec:motivation}

Reliable autonomous driving must jointly address three coupled challenges: learning efficiency, decision safety, and perceptual adaptability.

\textbf{Learning} Pure RL needs a lot of data and becomes unstable when reward signals are rare. IL uses data well but still drifts under covariate shift, and detailed rewards made by people often fail to match expert goals in complex traffic.

\textbf{Safety and Planning} Reactive policies are fast to respond, but cannot plan very far ahead. Generative planners can guess future events, but they often cannot run in real time or give strong safety guarantees, so there is still a gap between long-term planning and moment-to-moment control.

\textbf{Perception} Standard vision encoders treat almost all regions of an image uniformly and struggle to leverage contextual information effectively. In driving, however, attention mechanisms should be dynamic: prioritizing near-field lane coherence to ensure lateral stability at high speeds, while intensifying focus on immediate proximity when potential hazards are detected.

These limitations highlight the need for a unified framework that connects stable imitation with exploratory reinforcement learning.

\subsection{Contributions}
\label{sec:contributions}

We introduce a IRL–DAL, a cohesive framework that confronts the above challenges through three complementary components:

\textbf{(1) Hybrid IL–IRL–RL Training} 
A structured pipeline that integrates BC for initialization with PPO fine‑tuning under a hybrid reward combining sparse environment feedback and dense GAIL‑based intent rewards. This ensures stability, sample efficiency, and policy alignment with expert intent.

\textbf{(2) Diffusion Planner for On‑Demand Safety} 
A conditional diffusion model serves as a short‑horizon, risk‑aware planner, activated only in uncertain or high‑risk states. It generates candidate trajectories optimized by an energy‑based objective penalizing collisions and abrupt control variations, allowing the main policy to internalize safer behaviors via planner feedback.

\textbf{(3) Learnable Adaptive Mask (LAM)} 
A lightweight perception module that dynamically modulates spatial attention based on vehicle kinematics and LiDAR proximity. The mask directs the visual encoder toward context-critical regions—amplifying lower-field road features at high speed for precise lane keeping and highlighting proximate surroundings near hazards—achieving interpretable and efficient attention allocation without heavy self‑attention overhead.
\medskip

\begin{figure*}[t]
	\centering
	\includegraphics[width=\textwidth]{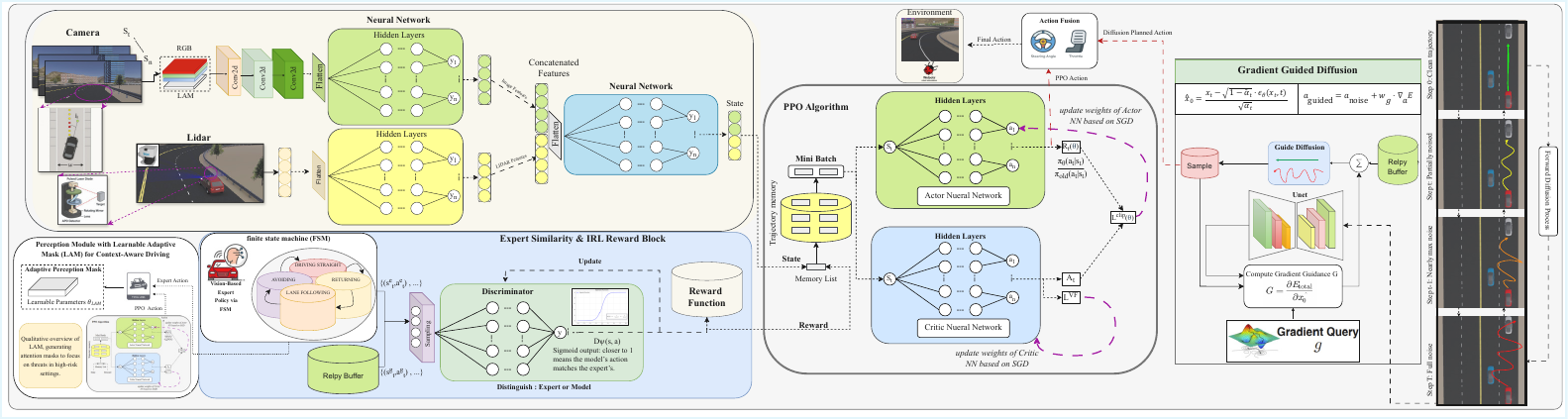}
	\caption{Overview of the IRL-DAL architecture. The training process unfolds in two phases: the policy is first initialized by Behavioral Cloning (BC) and then fine-tuned with Proximal Policy Optimization (PPO) using a hybrid reward $r_{\text{total}}$. A LAM enhances state-aware perception. During RL rollouts, the Diffusion-based Adaptive Lookahead (DAL) planner serves as a safety supervisor, correcting unsafe PPO actions through energy-guided sampling so that only safe experiences are stored in the replay buffer $\mathcal{D}_{\text{PPO}}$.}
	\label{fig:architecture_overview}
\end{figure*}

\section{Problem formulation}
\label{sec:Problem formulation}

The paper first formalizes the autonomous driving task as a partially observable Markov decision process POMDP defined by the tuple: 

\[
(\mathcal{S}, \mathcal{A}, \mathcal{T}, \mathcal{R}, \mathcal{O}, \mathcal{Z}, \gamma)
\]
which captures the agent partial observability and reliance on noisy sensor data.

State Space ($\mathcal{S}$)
The latent true state includes the ego car position, orientation, and speed. It also includes nearby moving objects, the road layout. The policy cannot see this state directly.

Action Space ($\mathcal{A}$)
Each action is a continuous two-dimensional vector:
\[
a_t = [\, \text{steering},\, \text{speed}\,]
\]

which is later mapped by a low-level controller into executable control commands. This design provides smooth, normalized continuous control.

Transition Function ($\mathcal{T}$)
The stochastic dynamics:
\[
\mathcal{T}(s_{t+1} \mid s_t, a_t)
\]
defines how the true state evolves given the action. 

Observation Space ($\mathcal{O}$)
The multimodal observation at each time step is given by:
\[
o_t = \{ I_t, L_t, K_t \}
\]
Where:
\begin{itemize}
	\item \textbf{Camera Image} ($I_t \in \mathbb{R}^{H \times W \times 3}$): front-facing RGB frame.
	\item \textbf{LiDAR Scan} ($L_t \in \mathbb{R}^{180}$): range readings from a forward LiDAR.
	\item \textbf{Vehicle Kinematics} ($K_t$): normalized ego-vehicle speed.
\end{itemize}
The LAM is generated internally based on $K_t$ and $L_t$. Thus, they belong to the perception module rather than the raw observation space.

Observation Function ($\mathcal{Z}$)
The conditional distribution:
\[
\mathcal{Z}(o_t \mid s_{t+1}, a_t)
\]
Models the sensing process that generates observations.

Reward Function ($\mathcal{R}$)
Since the agent perceives the environment only through observations, the reward depends on both $o_t$ and $a_t$. IRL-DAL uses a hybrid reward. It combines an environment-defined term with an intrinsic term learned by the IRL discriminator.
\[
r_t(o_t, a_t) = (1 - w_{\text{IRL}})\, r_{\text{env}}(o_t, a_t) + w_{\text{IRL}}\, r_{\text{IRL}}(o_t, a_t)
\]
where $w_{\text{IRL}}$ is a phase-dependent weight used during the mixed training phase, the influence of the learned reward throughout training.  Both terms depend solely on observable quantities, ensuring consistency under partial observability.

Discount Factor ($\gamma$)
A scalar $\gamma$ that balances short-term rewards with long-term performance.

Objective
The agent aims to learn a stochastic policy $\pi_\theta(a_t \mid o_t)$, parameterized by $\theta$, that maximizes the expected discounted return:
\[
\mathbb{E}_{\pi_\theta,\, \mathcal{T},\, \mathcal{Z}} 
\left[
\sum_{t=0}^{\infty} \gamma^t r_t
\right]
\]

\section{Methodology}
\label{sec:methodology}
Figure~\ref{fig:architecture_overview} provides an overview of the entire architecture. Our proposed framework integrates safety, stability, and expert-like decision-making within a single autonomous driving system. As illustrated in Figure~\ref{fig:architecture_overview}, the architecture combines four interacting components that collectively enable robust and adaptive behavior: (1) context-aware perception via the LAM, (2) FSM-aware structured replay buffers, (3) PPO fine-tuning with a hybrid reward, and (4) diffusion-based safety supervision with experience correction.

\begin{figure*}[t]
	\centering
	\includegraphics[width=1.0\textwidth]{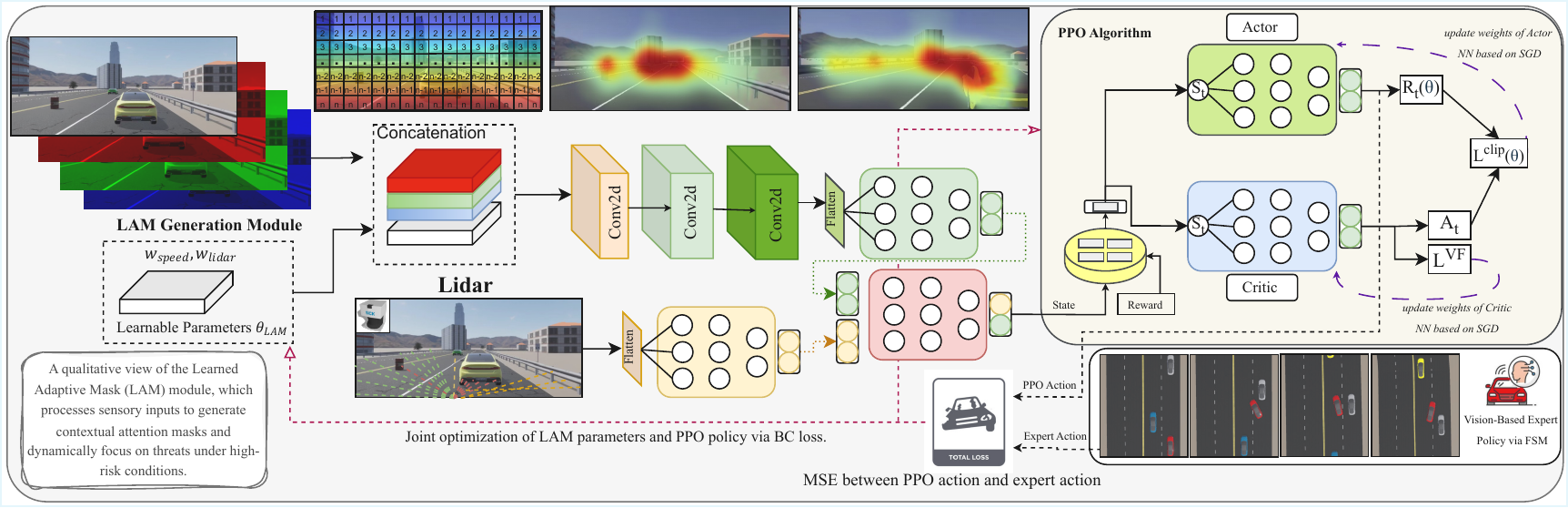}
	\caption{Architecture of the LAM and its integration with the PPO policy. Normalized speed $v_t^{\text{norm}}$ and hazard level $h_t$ modulate a vertical gradient mask via learnable parameters $\alpha_{\text{speed}}$ and $\alpha_{\text{lidar}}$. The resulting mask is concatenated with the RGB image to form a 4-channel input, enabling context-aware visual processing. LAM is trained end-to-end via BC gradients, allowing the agent to discover adaptive attention patterns that prioritize safety-critical regions.}
	\label{fig:lam_ppo_integration}
\end{figure*}
\subsection{Perception Module}
\label{sec:lam}

The perception module uses a learnable attention mechanism that dynamically focuses on the driving situation. This adaptive mask helps the agent build a compact but informative state representation $s_t$ from high-dimensional, multi-sensor inputs. The quality of the driving policy strongly depends on how well it sees and encodes the environment. Standard end-to-end methods often treat all image pixels equally, wasting model capacity. To fix this, we introduce the LAM, a small differentiable module that adds safety-related prior knowledge to the visual input using top-down attention.

Figure~\ref{fig:lam_ppo_integration} shows the LAM architecture and how it connects to the PPO policy. The module takes two inputs: the current vehicle speed $v_t$ and the minimum LiDAR distance $d_{\min,t} = \min(\min(l_t), d_{\max})$, where $l_t \in \mathbb{R}^{180}$ is the raw LiDAR range vector, clipped at the maximum range $d_{\max}$. These signals are then scaled to the interval $[0, 1]$.

\begin{align}
	v_t^{\text{norm}} &= \operatorname{clamp}\left( \frac{v_t}{v_{\max}}, 0, 1 \right) \label{eq:speed_norm} \\
	h_t &= \operatorname{clamp}\left( \frac{d_{\text{safe}} - d_{\min,t}}{d_{\text{safe}}}, 0, 1 \right) \label{eq:hazard_level}
\end{align}
where $v_{\max}$ and $d_{\text{safe}}$ are predefined maximum speed and safety distance thresholds, respectively. LAM computes context-dependent modulation factors using two learnable scalar parameters $\alpha_{\text{speed}}, \alpha_{\text{lidar}} \in \mathbb{R}$. They are initialized to $0.5$:
\begin{equation}
	f_{\text{speed}} = 1 + \alpha_{\text{speed}} \cdot v_t^{\text{norm}}, \quad
	f_{\text{hazard}} = 1 + \alpha_{\text{lidar}} \cdot h_t
	\label{eq:modulation_factors}
\end{equation}

These factors scale a base lower-bound intensity weight $w_{\text{base, lower}} = 1.0$, while the upper-bound weight is fixed at $w_{\text{base, upper}} = 0.0$. The resulting dynamic lower intensity is:

\begin{equation}
	w_{\text{lower}} = w_{\text{base, lower}} \cdot f_{\text{speed}} \cdot f_{\text{hazard}}
	\label{eq:lower_weight}
\end{equation}

A smooth vertical gradient mask is then constructed for each row $y \in [0, H-1]$ of the input image:

\begin{equation}
	M_t(y) = w_{\text{base, upper}} + (w_{\text{lower}} - w_{\text{base, upper}}) \cdot \frac{y}{H-1}
	\label{eq:mask_gradient}
\end{equation}

To ensure numerical stability and bounded output, the mask is normalized:
\begin{equation}
	\hat{M}_t = \frac{M_t}{\max(M_t) + \varepsilon}, \quad \varepsilon = 10^{-6}
	\label{eq:mask_norm}
\end{equation}
yielding $\hat{M}_t \in \mathbb{R}^{1 \times H \times W \times 1}$.

The normalized RGB image $I_t \in \mathbb{R}^{H \times W \times 3}$ (scaled to $[0,1]$) is concatenated channel-wise with $\hat{M}_t$ to form a 4-channel input tensor:

\begin{equation}
	I'_t = \operatorname{concat}(I_t / 255.0, \hat{M}_t) \in \mathbb{R}^{H \times W \times 4}
	\label{eq:4channel_input}
\end{equation}

This augmented input is processed by a shared convolutional backbone, which also fuses embedded LiDAR features before feeding into the actor and critic heads.

The LAM parameters $\theta_{\text{LAM}} = \{\alpha_{\text{speed}}, \alpha_{\text{lidar}}\}$ are optimized end-to-end during BC alongside the policy using the Adam optimizer with learning rate $\eta_{\text{BC}}$ and L2 regularization applied selectively to policy weights:

\begin{equation}
	\mathcal{L}_{\text{BC}} = \frac{1}{B} \sum_{i=1}^B \left\| \pi_{\theta}(s_t^{(i)}) - a_{\text{expert}}^{(i)} \right\|^2 + \lambda_{\text{L2}} \sum_{p \in \theta_{\text{policy}}} \|p\|^2
	\label{eq:bc_loss_with_l2}
\end{equation}

where gradients flow through the 4-channel observation to update $\theta_{\text{LAM}}$. Training includes gradient clipping (max norm $G_{\max}$) and learning rate scheduling via plateau detection.

As shown in Figure~\ref{fig:lam_ppo_integration}, the learned masks adapt dynamically to the driving context. Rather than shifting the geometric center of attention, the mechanism modulates the intensity of the spatial gradient. At high speeds, the mask amplitude significantly increases (via $f_{\text{speed}}$), which strengthens the feature extraction across the entire driveable area, effectively expanding the usable visual range while maintaining a strong prior on the immediate lane path. Similarly, when proximity to obstacles is detected ($d_{\min,t}$), the hazard factor ($f_{\text{hazard}}$) further amplifies the mask intensity. This ensures that the network receives a sharper, high-contrast signal of the immediate surroundings for precise collision avoidance, boosting safety without relying on hand-crafted heuristic rules.

\subsection{Multi-Phase Learning Curriculum}
\label{sec:learning_curriculum}

Policy learning happens in two main phases. First, BC gives the agent a stable and safe starting point by training the policy network on expert demonstrations. This warm start reduces risky, high-variance exploration and ensures the agent begins with reasonable behavior. Then, the policy is improved with PPO~\cite{schulman2017proximal}, which allows controlled on-policy exploration and lets the agent go beyond simple imitation.

To keep PPO updates aligned with expert behavior, an adversarial IRL module based on GAIL~\cite{ho2016generative} is added. This module provides dense, behavior-focused reward signals. The hybrid reward is defined as:

\begin{equation}
	r_t = w_{\text{env}} r_{\text{env}}(s_t, a_t) + w_{\text{irl}} r_{\text{irl}}(s_t, a_t)
	\label{eq:hybrid_reward}
\end{equation}

where $r_{\text{env}}(s_t, a_t)$ denotes the environment reward and $w_{\text{env}} + w_{\text{irl}} = 1$. The IRL reward is:
\begin{equation}
	r_{\text{irl}}(s_t, a_t) = -\log(1 - D_\psi(s_t, a_t) + \varepsilon)
\end{equation}
it is bounded within a predefined range.

The curriculum begins with an FSM-partitioned expert dataset $D_{\text{expert}}$, enabling balanced sampling across driving modes. During the imitation phase, both policy and diffusion planner are trained via BC with interval $T_{\text{BC}}^{\text{init}}$. In the mixed phase, PPO updates use hybrid rewards, with discriminator trained every $T_{\text{disc}}$ steps and BC continued adaptively every $T_{\text{BC}}^{\text{mixed}}$ steps. Safety interventions, which are triggered when $d_{\min,t} < d_{\text{safe}}$ or $|d_{\text{lane},t}| > e_{\text{lane}}$, are corrected and stored with a metadata flag indicating diffusion-based intervention. This progressive design, which moves from imitation to hybrid RL and then to safety distillation, leads to robust, safe, and efficient learning.

\subsection{Expert Data Generation via FSM-Aware Experience Replay}
\label{sec:fsm_data}

\begin{figure}[H]
	\centering
	\includegraphics[width=1.0\columnwidth]{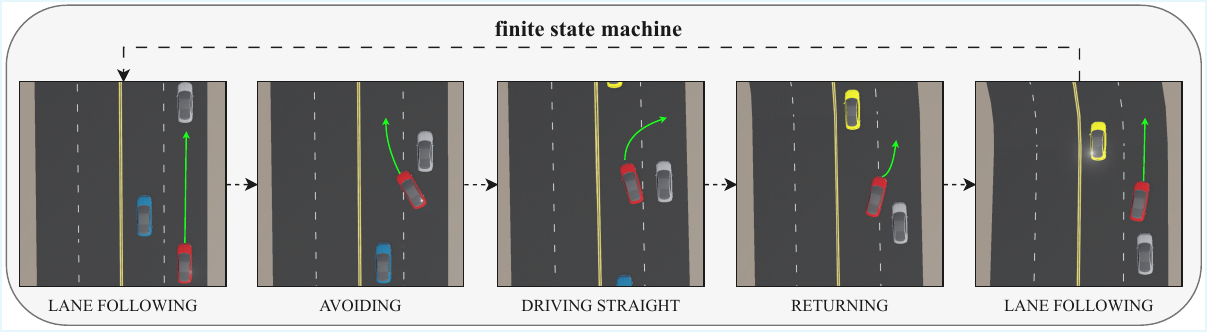}
	\caption{The FSM expert policy in action. It switches smoothly between the modes (Lane Following, Obstacle Avoidance, Driving Straight, Returning) using sensor-based transition rules. The FSM-aware experience replay stores each sample in its matching state buffer, which gives balanced exposure to both normal and risky driving situations.}
	\label{fig:fsm}
\end{figure}

A key part of the framework is a reliable and high-quality source of expert demonstrations. Instead of dealing with the noise and inconsistency of real human driving logs, an expert policy is built from scratch using a deterministic FSM policy $\pi^*$. As shown in Figure~\ref{fig:fsm}, this FSM controller moves smoothly between four behavior modes: Lane Following, Obstacle Avoidance, Driving Straight, and Returning, with transitions defined by simple sensor-based rules that describe the current driving situation. What sets this approach apart is the FSM-aware experience replay strategy. Each collected transition $(o_t, a_t^*, s_{\text{FSM},t})$ is stored in a separate buffer $D_s$ that corresponds to its active FSM state $s$. The full expert dataset is then formed as follows:

\[
D_{\text{expert}} = \bigcup_{s} D_s
\]

This clear, state-based structure helps solve a common problem in autonomous driving datasets, where rare but important events are often underrepresented, such as passing through a narrow gap or recovering from a strong lane drift. By sampling mini-batches evenly across FSM states, the training process sees a balanced mix of normal cruising and challenging edge cases, instead of being dominated by simple highway driving.

\subsubsection{Phase 1: Foundational Pre-training}
The training process starts with a supervised warm up over $N_{\text{imitation}}$ timesteps, using only $D_{\text{expert}}$ to build a safe and reliable behavioral base. This phase protects the agent from unstable and unsafe exploration that often appears at the start of RL. Under BC, the policy $\pi_\theta$ is trained on compact embeddings $s_t = \phi(o_t)$ from the perception module. It learns to match the expert actions using a mean-squared error loss. Balanced sampling from the FSM-partitioned buffers ensures that all driving modes are well represented during training:

\begin{equation}
	\mathcal{L}_{\text{BC}}(\theta) = 
	\mathbb{E}_{(o_t, a_t^*) \sim D_{\text{expert}}^{\text{balanced}}} \left[ \|\pi_\theta(s_t) - a_t^*\|^2 \right] + \lambda_{\text{L2}} \sum_{p \in \theta_{\text{policy}}} \|p\|^2
	\label{eq:bc_loss}
\end{equation}
This loss keeps the policy close to expert behavior long before any reward signal is used.
Diffusion planner training runs concurrently with BC. A conditional 1D U-Net diffusion planner is trained to generate smooth and feasible motion sequences. It is trained on consecutive expert action chunks $\{a_t^*, \dots, a_{t+H-1}^*\}$ taken from $D_{\text{expert}}$ and optimized with the standard DDPM denoising objective~\cite{ho2020denoising}. In this way, the planner learns the smooth and physically consistent control patterns of the FSM expert and is prepared to act later as a reliable safety net.

\subsubsection{Phase 2: Online Fine-tuning with Adversarial Reinforcement Learning}

After a strong base is built, the system runs for $N_{\text{mixed}}$ timesteps of online refinement with PPO. During this phase, the policy is improved through interaction with the environment, using a hybrid reward that combines clear task feedback with imitation-based signals. Under GAIL, a discriminator $D_\psi$ is trained to go beyond simple action matching. It learns to tell the difference between policy rollouts $(s_t, a_t)$ and true expert pairs $(s_t^*, a_t^*)$ using a binary cross-entropy loss:

\begin{equation}
	\begin{aligned}
		\mathcal{L}_{\text{Disc}}(\psi) =& - \mathbb{E}_{(s_t, a_t) \sim \pi_\theta} [\log(1 - D_\psi(s_t, a_t))] \\
		& - \mathbb{E}_{(s_t^*, a_t^*) \sim D_{\text{expert}}} [\log(D_\psi(s_t^*, a_t^*))]
	\end{aligned}
	\label{eq:discriminator_loss}
\end{equation}

Once training is complete, the discriminator provides a dense and continuous imitation reward:

\begin{equation}
	r_{\text{irl}}(s_t, a_t) = -\log(1 - D_\psi(s_t, a_t) + \varepsilon)
	\label{eq:gail_reward}
\end{equation}

Balanced sampling across FSM states ensures that the discriminator sees the full range of expert behaviors.

In the hybrid reward formulation, the final PPO reward is designed to balance task completion with expert-level behavior.

\begin{equation}
	r_t(s_t, a_t) = w_{\text{env}} r_{\text{env}}(s_t, a_t) + w_{\text{irl}} r_{\text{irl}}(s_t, a_t)
	\label{eq:hybrid_reward}
\end{equation}
where $w_{\text{env}} + w_{\text{irl}} = 1$. This combination helps the agent achieve high-level goals, such as steady progress and zero collisions, while also matching the smooth, anticipatory style of the FSM expert.

\subsection{Safety Assurance via Guided Diffusion and Experience Correction}
\label{sec:safety_assurance}

Finally, a diffusion-based planner is added as a safety-focused backup system. The DAL module is activated only in high-risk situations and quickly generates short, feasible paths that keep the vehicle safe. At the same time, it improves learning by storing safe, corrected experiences in the policy memory, so the agent becomes better over time without reinforcing unsafe behaviors. Together, these parts (perception, curriculum, expert replay, and safety) form an integrated pipeline that combines fast reactions with careful planning and leads to behavior that is effective, safe, and close to human driving.

Even when the PPO policy learns complex driving patterns, it is still hard to stay safe in rare or unexpected situations. To reduce these risks, the framework uses two safety components:

\begin{enumerate}
	\item An on-demand, energy-guided diffusion planner that is activated to generate safe actions when the situation becomes dangerous.
	\item An experience correction mechanism that sends these safe corrections back to the main policy so that temporary fixes become lasting improvements.
\end{enumerate}

\subsubsection{On-Demand, Energy-Guided Trajectory Generation}
The diffusion planner trained in the first phase now acts as a safety-critical motion generator. It stays inactive most of the time to save computation and is only turned on in high-risk states. A state is treated as high-risk when the closest LiDAR reading becomes too small or the vehicle moves far away from the lane center:

\begin{equation}
	d_{\min,t} < d_{\text{trigger}} \quad \vee \quad |d_{\text{lane},t}| > e_{\text{lane}}
	\label{eq:dal_trigger}
\end{equation}
letting the PPO policy drive freely under normal conditions.

When DAL is called, it generates a short safe trajectory $\mathcal{A} = {a_t, \dots, a_{t+H-1}}$ based on the current context embedding. This is done with a guided reverse diffusion process. At each denoising step $k$, the U-Net predicts a cleaner trajectory $\hat{\mathcal{A}}^0_k$, which is then moved toward safer behavior by following the gradient of a composite energy function $E(\mathcal{A}, o_t)$:
\begin{equation}
	\tilde{\mathcal{A}}^0_k = \hat{\mathcal{A}}^0_k - w_g \cdot \frac{\nabla E(\hat{\mathcal{A}}^0_k, o_t)}{\|\nabla E\| + \varepsilon}
	\label{eq:dal_guidance}
\end{equation}
where $w_g$ determines how strongly the planner pushes trajectories toward safe regions during sampling, and $\tilde{\mathcal{A}}^{0}_{k}$ denotes the resulting safety-guided trajectory obtained after applying the energy gradient.

The energy $E$ is a weighted sum of five simple terms that together enforce lane keeping, obstacle clearance, smooth control, stability, and an optional expert-alignment term.
\begin{equation}
	E = E_{\text{lane}} + E_{\text{lidar}} + E_{\text{jerk}} + E_{\text{stability}} + E_{\text{expert}}
	\label{eq:dal_energy}
\end{equation}

Lane Adherence — $E_{\text{lane}}$: 
This term penalizes deviation from the lane center. Instead of computationally expensive forward modeling, the energy term utilizes the current lateral error state distance ($d_{\text{lat}}$) to strictly guide the diffusion sampling process toward immediate correction.

\begin{equation}
	E_{\text{lane}} = w_{\text{lane}} \left( \frac{d_{\text{lane},t}}{s_{\text{lane}}} \right)^2
\end{equation}
with a risk-adaptive weight $w_{\text{lane}}$ that increases under higher hazard levels:
\[
w_{\text{lane}} = w_{\text{base}}^{\text{lane}} (1 + \alpha_{\text{hazard}} h_t)
\]
where $w_{\text{base}}^{\text{lane}}$ is the nominal weighting coefficient for lane keeping, 
$\alpha_{\text{hazard}}$ controls how aggressively the weight increases under hazardous conditions, 
and $h_t$ denotes the instantaneous hazard indicator.

Obstacle Avoidance — $E_{\text{lidar}}$: 
To keep a safe distance from obstacles, trajectories generated while the vehicle is in close proximity to obstacles are penalized based on the current sensor state ($d_{\min,t}$).
\begin{equation}
	E_{\text{lidar}} = w_{\text{lidar}} \cdot \max\left(0, \frac{d_{\text{safe}}^{\text{plan}} - d_{\min,t}}{s_{\text{lidar}}} \right)^2
\end{equation}
where the weight $w_{\text{lidar}}$ increases with the hazard level, according to
$w_{\text{lidar}} = w_{\text{base}}^{\text{lidar}} (1 + h_t)$.

Control Smoothness — $E_{\text{jerk}}$:  
To avoid jerky or hard-to-drive behavior, large changes between consecutive actions are penalized:
\begin{equation}
	E_{\text{jerk}} = 
	w_{\text{jerk}} \cdot \frac{1}{H-1} 
	\sum_{i=1}^{H-1} \|a_i - a_{i-1}\|^2
\end{equation}
where $w_{\text{jerk}}$ is the weighting coefficient for smoothness and $H$ is the trajectory horizon.

Stability — $E_{\text{stability}}$:  
Steady, centered control is encouraged by penalizing large steering or speed deviations from a neutral reference:
\begin{equation}
	E_{\text{stability}} = 
	w_{\text{stab}} \cdot \frac{1}{H} 
	\sum_{i=0}^{H-1} 
	\left( \text{steer}_i^2 + (\text{speed}_i - v_{\text{ref}})^2 \right)
\end{equation}
where $w_{\text{stab}}$ is the stability weight and $v_{\text{ref}}$ denotes the target reference speed.

Expert Alignment — $E_{\text{expert}}$ (optional):  
When an expert reference is available, an additional term keeps the trajectory close to the expert demonstration:
\begin{equation}
	E_{\text{expert}} = 
	w_{\text{exp}} \cdot \frac{1}{H} 
	\sum_{i=0}^{H-1} \|a_i - a_i^{\text{expert}}\|^2
\end{equation}
where $w_{\text{exp}}$ controls the strength of expert-matching and $a_i^{\text{expert}}$ is the corresponding expert action.

The hazard level $h_t$ adapts all risk-sensitive weights:
\begin{equation}
	h_t = 
	\operatorname{clamp}\!\left(
	1 - \tanh\!\left(\frac{d_{\min,t}}{s_h}\right), 0, 1
	\right)
\end{equation}
where $d_{\min,t}$ is the minimum LiDAR distance at time $t$, $s_h$ is a scaling factor determining sensitivity to obstacles, and the clamp limits the value to the range $[0,1]$.

Conditioning is performed using a compact context vector 
$c_t \in \mathbb{R}^{64}$, which encodes current hazard level, lateral deviation, speed, steering, and LiDAR statistics. Each element is normalized, and unused entries are padded for dimensional consistency.

\subsubsection{Action Blending and Execution}

Only the first action $a_t^{\text{DAL}}$ from the refined trajectory is used. During the mixed phase, this action is smoothly blended with the PPO action using a dynamic blend weight:
\begin{equation}
	w_b = 
	\begin{cases}
		1.0, & d_{\min,t} < d_{\text{critical}} \\
		w_{\text{base}}^{\text{blend}} + w_{\text{scale}}^{\text{blend}} h_{\text{blend}} & \text{otherwise}
	\end{cases}
	\label{eq:blend_critical}
\end{equation}
where:
\begin{equation}
	h_{\text{blend}} = e^{-d_{\min,t}/s_1} + k_{\text{tanh}} \tanh(|d_{\text{lane},t}|/s_2)
	\label{eq:blend_hazard}
\end{equation}
and $w_b$ is temporally smoothed via EMA. The final executed action is:
\begin{equation}
	a_t^{\text{final}} = w_b a_t^{\text{DAL}} + (1 - w_b) a_t^{\text{PPO}}
	\label{eq:final_blend}
\end{equation}

\subsubsection{Experience Correction}
\begin{figure}[H]
	\centering
	\includegraphics[scale=0.3]{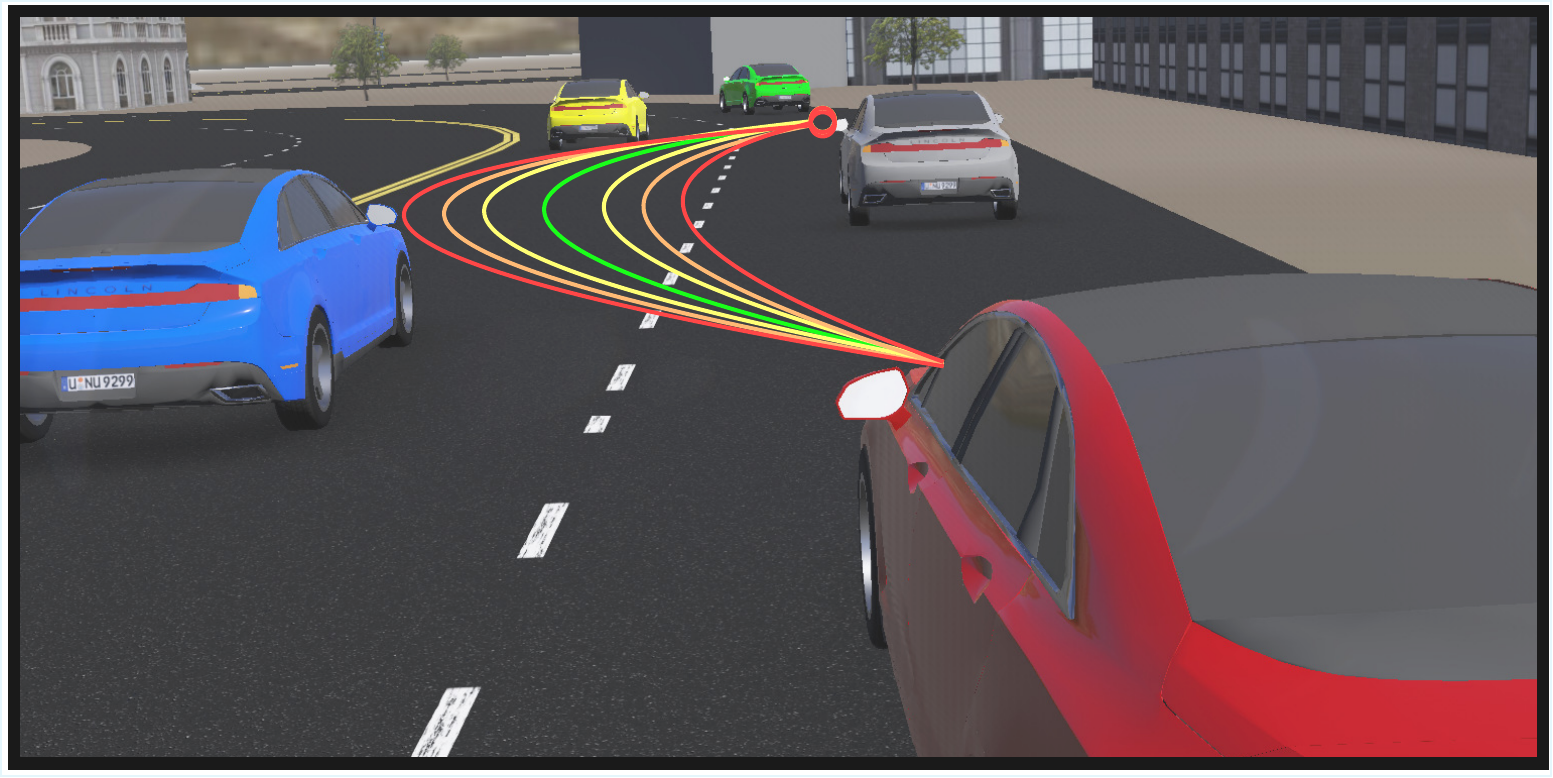}
	\caption{The safety pipeline in action. DAL is activated by high-risk signals, generates energy-guided safe trajectories, and blends the first action with the PPO output. The corrected experiences are stored in the replay buffer with markers, The diffusion planner acts as an active shield, allowing the agent to safely collect more high-quality expert data even in challenging scenarios.}
	\label{fig:safety_pipeline}
\end{figure}

The executed transition is stored in the PPO replay buffer with a correction marker, and any unsafe PPO action is replaced by the safe blended output to prevent accidents (Runtime Shielding). However, the experience stored in the replay buffer remains anchored to the FSM expert action. This ensures the policy learns stable rule-based behaviors while the diffusion layer prevents premature episode termination, as illustrated in Fig.~\ref{fig:safety_pipeline}.

\subsection{Safety-Aware Experience Correction (SAEC)}
\label{sec:experience_correction}

To ensure robust training without premature episode termination, the SAEC mechanism employs the diffusion planner as a runtime safety shield. As described in Algorithm 1, the planner is activated when the agent enters a high-risk state, defined by:

\[
d_{\min,t} < d_{\text{trigger}} \vee |d_{\text{lane},t}| > e_{\text{lane}} \tag{27}
\]

where $d_{\min,t}$ is the closest LiDAR reading and $d_{\text{lane},t}$ measures the lateral deviation. Once triggered, DAL generates a safe action to prevent immediate collision. However, to ensure the policy converges to a stable kinematic behavior, we do not train on the corrective diffusion action itself. Instead, we utilize the concurrent Rule-Based Expert (FSM) to calculate the ideal ground-truth action ($a_{\text{Expert}}$) for that specific state.

The transition stored in the PPO replay buffer $D_{\text{PPO}}$ is therefore defined as:

\[
D_{\text{PPO}} \leftarrow D_{\text{PPO}} \cup (o_t, a_{\text{Expert}}, r_t, o_{t+1}) \tag{28}
\]

This mechanism provides two critical benefits:

\begin{itemize}
	\item \textbf{Runtime Survival:} The diffusion action is executed to physically prevent collisions, allowing the episode to continue beyond states that would normally cause termination.
	\item \textbf{Safe Expert Labeling:} By keeping the agent active in near-accident scenarios, the system allows the FSM expert to label these "edge cases" with correct recovery actions. The policy $\pi_{\theta}$ thus learns obstacle avoidance based on the FSM's consistent logic, while the diffusion layer acts solely as a guardrail during this learning process.
\end{itemize}
\subsection{Policy Optimization with Safety-Aware Curriculum}
\label{sec:policy_optimization}
The core of the system, the driving policy, is trained with a safety-first curriculum that combines the stable reliability of IL with the flexibility of RL. This combination leads to behavior that is not only skilled but also naturally cautious and robust. The full training procedure is summarized in Algorithm~\ref{alg:main_training_loop}.
\subsubsection{RL Backbone: PPO}
To improve the policy through interaction with the environment, PPO is used, an on-policy actor-critic method well suited to smooth, high-dimensional control. Its clipped objective keeps updates stable and prevents large policy changes that could disrupt training.
The full optimization objective is:
\begin{equation}
	L(\theta) = \hat{\mathbb{E}}_t \left[ L_t^{\text{CLIP}}(\theta) + c_1 L_t^{\text{VF}}(\phi) - c_2 S[\pi_\theta](s_t) \right]
	\label{eq:ppo_loss}
\end{equation}
where $L_t^{\text{CLIP}}$ is the clipped policy loss, $L_t^{\text{VF}}$ is the value function error, and $S[\pi_\theta]$ is an entropy term that encourages the policy to keep exploring. Raw observations are first converted into compact embeddings $s_t = \phi(o_t)$ by a shared perception backbone (Sec.~\ref{sec:lam}). After this step, the actor and critic each use their own lightweight MLP head.
\subsubsection{Multi-Phase Training Curriculum}
Training is carried out in two phases. It starts with an imitation warm up and then moves to a reward driven refinement phase.
\paragraph{Phase 1 — Imitation Pre-training:}  
Over the first $N_{\text{imitation}}$ timesteps, both the policy $\pi_\theta$ and diffusion planner are bootstrapped via BC on the FSM-aware expert dataset $D_{\text{expert}}$ (Sec.~\ref{sec:fsm_data}). Balanced sampling across driving modes guarantees exposure to rare but critical edge cases:
\begin{equation}
	\mathcal{L}_{\text{BC}} = \mathbb{E}_{(o_t, a_t^*) \sim D_{\text{expert}}^{\text{balanced}}} \left[ \| \pi_\theta(s_t) - a_t^* \|^2 \right] + \lambda_{\text{L2}} \|\theta_{\text{policy}}\|^2.
	\label{eq:bc_loss}
\end{equation}
This phase plants a \textbf{safe, expert-grade behavioral prior}, slashing unsafe exploration right from the start.
\paragraph{Phase 2 — IRL-PPO Fine-tuning with Hybrid Reward:}  
Once a solid foundation is set, the agent goes for $N_{\text{mixed}}$ timesteps, refining itself with PPO under a \textbf{hybrid reward} that fuses crisp environmental feedback with rich, learned imitation cues:
\begin{equation}
	r_t = w_{\text{env}} r_{\text{env}}(s_t, a_t) + w_{\text{irl}} r_{\text{irl}}(s_t, a_t)
	\label{eq:hybrid_reward_final}
\end{equation}
where $w_{\text{env}} + w_{\text{irl}} = 1$.

IRL Reward from GAIL discriminator:  
\[
r_{\text{irl}}(s_t, a_t) = -\log(1 - D_\psi(s_t, a_t) + \varepsilon), \quad r_{\text{irl}} \in [r_{\min}, r_{\max}]
\]

\begin{figure}[H]
	\centering
	\includegraphics[scale=1.1]{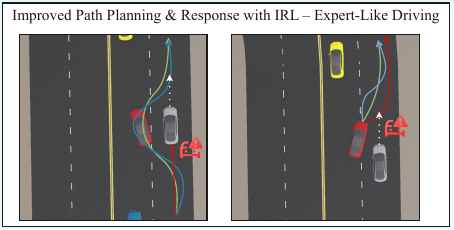}
	\caption{Impact of Hybrid Reward Shaping on Trajectory Smoothness. The blue trajectory illustrates the policy trained solely on the rule-based environment reward ($r_{\text{env}}$), resulting in oscillatory behavior and excessive lateral deviation. In contrast, the green trajectory incorporates the dense IRL signal ($r_{\text{IRL}}$), effectively regularizing the policy towards the expert’s kinematic profile. This results in smoother lane-change maneuvers (left) and tighter collision avoidance (right), correcting the overshooting tendencies observed in the baseline.}
	\label{fig:irl_trajectory_comparison}
\end{figure}

The environment reward $r_{\text{env}}$ is dense and FSM aware. It is adapted to the current driving mode and combines terms for precise lane centring, obstacle clearance, and sparse goal completion. A collision gives a large negative reward, a near miss gives a smaller penalty, and a minimum baseline keeps the reward positive in safe states, as illustrated in Fig.~\ref{fig:irl_trajectory_comparison}.

\subsection{Integrated Training Pipeline}
Algorithm~\ref{alg:main_training_loop} ties everything together: FSM-aware replay, BC, GAIL, PPO, DAL, and SAEC.

\begin{algorithm}[t]
	\caption{Hybrid IRL--DAL Training Framework}
	\label{alg:main_training_loop}
	\DontPrintSemicolon
	
	\textbf{Initialize:} $\pi_\theta$, $V_\phi$, $P_{\text{DAL}}$, $D_\psi$,
	$\mathcal{D}_{\text{buffer}} \leftarrow \emptyset$\;
	
	\textbf{Phase 1: Imitation (Warm-up \& Data Collection)}
	
	\For{$t \gets 0$ \KwTo $N_{\text{imitation}} - 1$} {
		Observe $o_t$, compute expert action $a_t^* \leftarrow \text{FSM}(o_t)$\;
		Execute $a_t^*$, observe $o_{t+1}$\;
		Store $(o_t, a_t^*)$ in $\mathcal{D}_{\text{buffer}}$\;
		\If{$t \bmod T_{\text{BC}} = 0$} {
			Update $\pi_\theta$ via BC on $\mathcal{D}_{\text{buffer}}$ (Eq.~\ref{eq:bc_loss})\;
		}
		\If{$t \bmod T_{\text{diffusion}} = 0$} {
			\textsc{TrainDiffusion}($\mathcal{D}_{\text{buffer}}$)\;
		}
	}
	
	\textbf{Phase 2: Mixed (RL + DAL Safety + Regularization)}
	
	\For{$t \gets N_{\text{imitation}}$ \KwTo $N_{\text{total}} - 1$} {
		Sample agent action $a_t^{\text{PPO}} \sim \pi_\theta(o_t)$\;
		Compute expert action $a_t^* \leftarrow \text{FSM}(o_t)$\;
		
		\uIf{$d_{\min,t} < d_{\text{trigger}}$ \textbf{ or } $|d_{\text{lane},t}| > e_{\text{lane}}$} {
			Sample $\mathcal{A} \sim P_{\text{DAL}}(o_t)$ with energy guidance\;
			$a_t^{\text{final}} \gets \text{blend}(a_t^{\text{DAL}}, a_t^{\text{PPO}})$\;
		}
		\Else {
			$a_t^{\text{final}} \gets a_t^{\text{PPO}}$\;
		}
		
		Execute $a_t^{\text{final}}$, observe $r_{\text{env}}, o_{t+1}$\;
		$r_{\text{irl}} \gets -\log(1 - D_\psi(o_t, a_t^{\text{final}}) + \varepsilon)$\;
		$r_t \gets w_{\text{env}} r_{\text{env}} + w_{\text{irl}} r_{\text{irl}}$\;
		
		Store $(o_t, a_t^{\text{final}}, a_t^*, r_t)$ in $\mathcal{D}_{\text{buffer}}$\;
		
		\If{$t \bmod T_{\text{disc}} = 0$} {
			\textsc{TrainDiscriminator}($\mathcal{D}_{\text{buffer}}$)\;
		}
		\If{$t \bmod T_{\text{sync}} = 0$} {
			\textsc{SyncFeatures}()\;
		}
		\If{\textsc{should\_run\_bc\_training}()} {
			\textsc{TrainBC}($\mathcal{D}_{\text{buffer}}$)\;
		}
		Update $\pi_\theta, V_\phi$ via PPO using collected rollouts\;
	}
\end{algorithm}

\begin{table*}[t]
	\centering
	\caption{Quantitative performance across architectural variants (10 seeds, mean $\pm$ std). Mean reward normalized to [0, 200]. Trajectory prediction metrics (ADE/FDE) from rollout evaluation. Arrows indicate improvement direction; \textbf{bold} denotes best.}
	\label{tab:metrics}
	\resizebox{\linewidth}{!}{
		\begin{tabular}{lccccccc}
			\toprule
			\textbf{Model} & \textbf{Mean Reward} $\uparrow$ & \textbf{Coll./1k Steps} $\downarrow$ & \textbf{Success (\%)} $\uparrow$ & \textbf{BC Loss ($\times 10^{-2}$)} $\downarrow$ & \textbf{Action Sim. (\%)} $\uparrow$ & \textbf{ADE (m)} $\downarrow$ & \textbf{FDE (m)} $\downarrow$ \\
			\midrule
			PPO + Uniform Sampling & 85.2 $\pm$ 4.1 & 0.63 $\pm$ 0.12 & 78.1 $\pm$ 3.2 & 17.1 $\pm$ 1.4 & 65.3 $\pm$ 4.1 & 5.25 $\pm$ 0.31 & 11.8 $\pm$ 0.65 \\
			+ FSM Replay & 120.4 $\pm$ 3.8 \textcolor{red}{(+41\%)} & 0.30 $\pm$ 0.08 & 88.4 $\pm$ 2.1 & 12.3 $\pm$ 1.1 & 75.1 $\pm$ 3.5 & 4.10 $\pm$ 0.27 & 9.5 $\pm$ 0.58 \\
			+ Diffusion Planner & 155.1 $\pm$ 3.2 \textcolor{red}{(+29\%)} & 0.15 $\pm$ 0.05 & 92.0 $\pm$ 1.8 & 13.0 $\pm$ 1.0 & 80.2 $\pm$ 3.0 & 3.15 $\pm$ 0.22 & 7.2 $\pm$ 0.49 \\
			\textbf{+ LAM + SAEC (Ours)} & \textbf{180.7 $\pm$ 2.9} \textcolor{red}{(+16\%)} & \textbf{0.05 $\pm$ 0.03} & \textbf{96.3 $\pm$ 1.2} & \textbf{7.4 $\pm$ 0.8} & \textbf{85.7 $\pm$ 2.4} & \textbf{2.45 $\pm$ 0.18} & \textbf{5.1 $\pm$ 0.41} \\
			\bottomrule
	\end{tabular}}
\end{table*}

\section{Simulations and Results}
\label{sec:Results}

\subsection{Experimental Setup}

All experiments are done in the Webots simulator. Webots provides realistic vehicle motion and flexible scene settings. The virtual city has multi-lane curved roads, moving obstacles, and different lighting conditions from bright day to dark evening. This setup makes the tests close to real urban driving. For reproducibility, all details of the perception backbone are given. It takes a dictionary-style observation that includes camera images and LiDAR scans.

\begin{enumerate}
	\item \textbf{Input Data:}
	The model takes two input streams. The visual stream is a 4-channel tensor $I'_t \in \mathbb{R}^{H \times W \times 4}$, formed from a downscaled RGB image and the LAM output. The auxiliary stream is a 1D LiDAR scan $L_t \in \mathbb{R}^{N_{\text{beams}}}$ that provides distance measurements in all directions.
	
	\item \textbf{Visual Encoder:}
	A three-layer convolutional network with kernel sizes 5, 3, and 3 processes the visual input. It uses ReLU activations and stride 2 for downsampling, and a final flatten layer produces a compact visual feature vector $z_{\text{vision}}$.
	
	\item \textbf{LiDAR Encoder:}
	The LiDAR vector is first normalized and then passed through a three-layer MLP with ReLU activations. This network produces a 32-dimensional embedding $z_{\text{lidar}}$ that captures the scene geometry.
	
	\item \textbf{Feature Fusion:}
	The two latents are concatenated and polished by a small fusion MLP to produce the final state $s_t \in \mathbb{R}^{512}$:
	\begin{equation}
		\bm{s}_t = \text{MLP}_{\text{fusion}}(\text{concat}[\bm{z}_{\text{vision}}, \bm{z}_{\text{lidar}}]).
		\label{eq:feature_fusion}
	\end{equation}
	This fused vector gives the policy a complete view of the scene, with fine local structure from LiDAR and global context from vision, and it is used in both the imitation warm-up and the later online RL training.
	
	\item \textbf{Environment and Episode Termination:}
	A custom Gym wrapper updates observations at 10 Hz. At each step, it provides an $H \times W \times 4$ visual tensor (RGB plus mask) and an $N_{\text{beams}}$-beam LiDAR scan. An episode ends either when a collision occurs ($d_{\min} < d_{\text{collision}}$) or when the goal is reached. A successful episode is one that reaches the goal without any violations.
	
	\item \textbf{Training Procedure:}
	Training followed the two-phase safety-aware curriculum shown in Algorithm~\ref{alg:main_training_loop}, for a total of $N_{\text{total}}$ timesteps. All transitions were stored in FSM-partitioned replay buffer, so that rare safety-related events are well represented. The main hyperparameters are given in Table~\ref{tab:params}.
\end{enumerate}

\begin{table}[t]
	\centering
	\caption{Training hyperparameters}
	\label{tab:params}
	\resizebox{0.85\columnwidth}{!}{
		\begin{tabular}{l|c}
			\toprule
			\textbf{Parameter} & \textbf{Value (from code)} \\ 
			\midrule
			PPO Steps / Batch / Epochs & 2048 / 64 / 10 \\
			Discount Factor ($\gamma$) / GAE-$\lambda$ & 0.99 / 0.95 \\
			BC / PPO Learning Rate & $3 \times 10^{-4}$ \\
			Diffusion Horizon ($H$) / Diffusion Steps ($T$) & 8 / 100 \\
			Total Buffer Capacity & 50{,}000 transitions \\
			IRL Reward Weight ($w_{\text{irl}}$) & 0.3 \\
			Environment Reward Weight ($w_{\text{env}}$) & 0.7 \\
			Imitation Phase Duration ($N_{\text{imitation}}$) & 20{,}000 steps \\
			Mixed Phase Duration ($N_{\text{mixed}}$) & 30{,}000 steps \\
			Total Training Steps ($N_{\text{total}}$) & 50{,}000 steps \\
			BC Update Interval (Imitation) & every 1{,}000 steps \\
			BC Update Interval (Mixed) & every 500 steps \\
			Diffusion Training Interval & every 500 steps \\
			Discriminator Update Interval ($T_{\text{disc}}$) & every 2{,}000 steps \\
			Feature Sync Interval ($T_{\text{sync}}$) & every 1{,}000 steps \\
			L2 Regularization ($\lambda_{\text{L2}}$) & $10^{-6}$ \\
			PPO Clip Range & 0.2 \\
			Entropy Coefficient ($c_2$) & 0.01 \\
			Value Loss Coefficient ($c_1$) & 0.5 \\
			Gradient Clip Norm ($G_{\max}$) & 0.5 \\
			Learning Rate Scheduler & Reduce on Plateau (patience=5) \\
			LiDAR Beams ($N_{\text{beams}}$) & 180 \\
			Image Size ($H \times W$) & $64 \times 64$ \\
			Collision Threshold ($d_{\text{collision}}$) & 1.0 m \\
			DAL Trigger: $d_{\text{trigger}}$ / $e_{\text{lane}}$ & 3.0 m / 120 px \\
			Critical Blend Threshold ($d_{\text{critical}}$) & 1.5 m \\
			Hazard Scale ($s_h$) & 3.0 \\
			Blend Hazard Scales ($s_1$, $s_2$, $k_{\text{tanh}}$) & 2.0 / 20 / 0.3 \\
			DAL Guidance Weight ($w_g$) & 0.1 \\
			Energy Weights ($w_{\text{jerk}}$, $w_{\text{stab}}$, $w_{\text{exp}}$) & 0.1 / 0.5 / 2.0 \\
			Lane Scale ($s_{\text{lane}}$) / LiDAR Scale ($s_{\text{lidar}}$) & 0.5 / 2.0 \\
			LAM Alphas Initial Value & 0.5 \\
			Optimizer & Adam ($\beta_1=0.9$, $\beta_2=0.999$) \\
			\bottomrule
	\end{tabular}}
\end{table}

\subsection{Component-wise Evaluation of the IRL–DAL Framework}

\begin{figure}[t]
	\centering
	\includegraphics[width=1\linewidth]{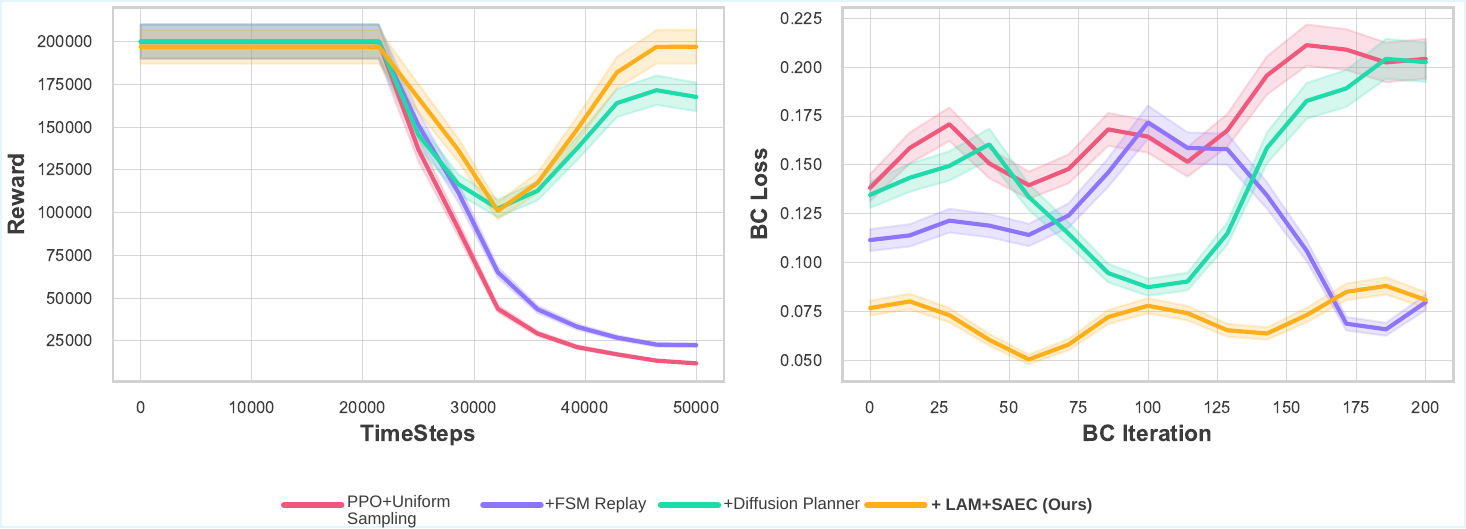}
	\caption{Training dynamics over 50k steps. After switching to mixed-mode at 20k steps, the full IRL–DAL model (\textit{Diff+Adapt+SAEC}) rockets ahead in reward and stability. BC loss stays pinned low, and DAL interventions fade to near-zero—proof of \textbf{adaptive distillation}. See Table~\ref{tab:metrics} for final numbers.}
	\label{fig:reward_bc}
\end{figure}

To measure the effect of each component, a clear ablation study was carried out. Four model variants were trained under the same conditions, adding components one by one to separate their individual contributions. The evaluated variants are as follows: The Baseline (PPO + Uniform Sampling) is the standard PPO algorithm with a uniform replay buffer. The second variant, + Structured Replay, builds upon the baseline model by incorporating FSM-aware replay, which aims to provide more coverage to safety-critical states. The third variant, + Generative Planner, includes the structured replay along with the diffusion-based safety planner, but without adaptive perception. Finally, the + LAM + SAEC (Full IRL-DAL) represents the full model, incorporating the adaptive mask, safety-aware experience correction, and all components enabled. These results are summarized in Table~\ref{tab:metrics} and visualized in Fig.~\ref{fig:reward_bc}, which display reward curves, BC alignment, and intervention decay.

\begin{figure}[t]
	\centering
	\includegraphics[width=1\linewidth]{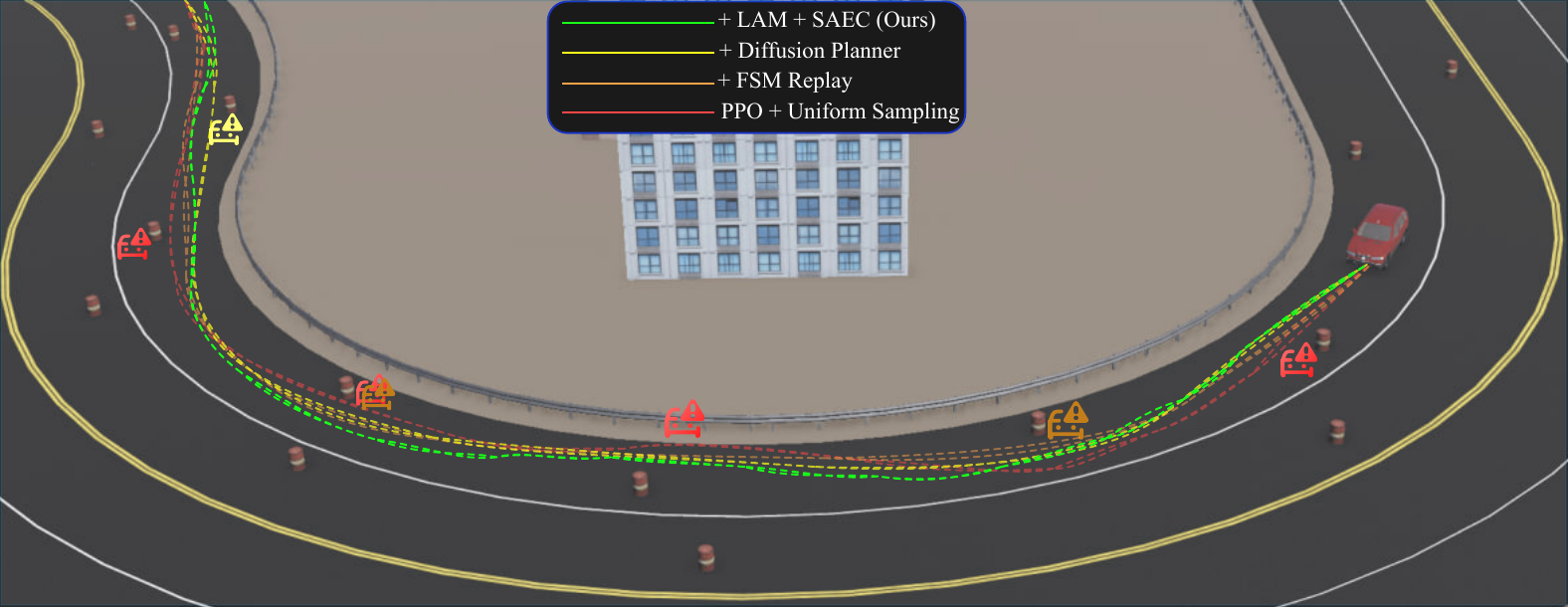} 
	\caption{Qualitative ablation study on a high-curvature turn scenario in Webots. The \textbf{Baseline PPO (Red)} fails to negotiate the curve, resulting in a collision. Adding \textbf{FSM Replay (Orange)} improves lane adherence but exhibits oscillatory behavior. The \textbf{Diffusion Planner (Yellow)} successfully ensures safety but deviates from the center. The full \textbf{IRL--DAL framework (Green)} demonstrates superior trajectory smoothness and precise lane centering, attributing to the LAM's focused attention on road boundaries.}
	\label{fig:trajectory_ablation}
\end{figure}

Structured replay and safety awareness have a key role in the results. Using only the FSM buffer increases the reward by 41\% and reduces collisions by 52\% (from 0.63 to 0.30 per 1k steps), which shows that rare events need special attention. Adding DAL increases the reward by another 29\% and again cuts collisions by half (from 0.30 to 0.15), so it not only reacts to danger but also helps prevent it (Fig.~\ref{fig:trajectory_ablation}). With the full model, including adaptive perception (LAM) and SAEC, collisions drop to 0.05, which is a 67\% reduction compared to the previous step and 12.6 times better than the baseline. At the same time, BC loss stays low and action similarity to the expert increases, so expert-like behavior is reached without sacrificing stability.

\section{Conclusion}
\label{sec:conclusion}

This work introduced IRL-DAL, a unified framework for safe and adaptive autonomous driving that combines inverse reinforcement learning, diffusion planning, and on-policy control. The method links three key ideas in a single loop: hybrid IL-IRL-RL training, a diffusion-based safety planner, and a LAM for perception. Together, these parts let the agent learn an expert-like driving policy that remains stable under online interaction and reacts safely in high-risk situations.

The framework was evaluated in the Webots simulator using a two-phase curriculum. In the first phase, an FSM expert and behavioral cloning provided a safe and reliable starting point for both the policy and the diffusion planner. In the second phase, PPO with a hybrid reward and GAIL-based IRL signals refined the behavior while DAL and SAEC enforced safety and turned interventions into useful training data. The final agent reached a 96\% success rate and reduced collisions to 0.05 per 1k steps. Ablation studies showed clear gains from each component: FSM-aware replay improved coverage of rare events, the diffusion planner reduced further failures, and LAM plus SAEC delivered the larger safety and performance gains.

\bibliographystyle{IEEEtran}
\bibliography{References}

@article{crosato2024social,
  title={Social Interaction-Aware Dynamical Models and Decision-Making for Autonomous Vehicles},
  author={Crosato, Luca and Tian, Kai and Shum, Hubert PH and Ho, Edmond SL and Wang, Yafei and Wei, Chongfeng},
  journal={Advanced Intelligent Systems},
  volume={6},
  number={3},
  pages={2300575},
  year={2024},
  publisher={Wiley Online Library}
}

@article{kiran2021deep,
  title     = {Deep reinforcement learning for autonomous driving: A survey},
  author    = {Kiran, B. Ravi and Sobh, Ibrahim and Talpaert, Victor and Mannion, Patrick and Sallab, Ahmad El and Yogamani, Senthil and P{\'e}rez, Patrick},
  journal   = {IEEE Transactions on Intelligent Transportation Systems},
  volume    = {23},
  number    = {6},
  pages     = {4909--4926},
  year      = {2021}
}

@inproceedings{ross2011reduction,
  title={A reduction of imitation learning and structured prediction to no-regret online learning},
  author={Ross, St{\'e}phane and Gordon, Geoffrey and Bagnell, J Andrew},
  booktitle={Proceedings of the fourteenth international conference on artificial intelligence and statistics},
  pages={627--635},
  year={2011}
}

@article{cui2023safe,
  title={Safe and Human-Like Trajectory Planning of Self-Driving Cars: A Constraint Imitative Method},
  author={Cui, Mingyang and Hu, Yingbai and Xu, Shaobing and Wang, Jianqiang and Bing, Zhenshan and Li, Boqi and Knoll, Alois},
  journal={Advanced Intelligent Systems},
  volume={5},
  number={10},
  pages={2300269},
  year={2023},
  publisher={Wiley Online Library}
}

@article{li2022driver,
  title={Driver behavioral cloning for route following in autonomous vehicles using task knowledge distillation},
  author={Li, Guofa and Ji, Zefeng and Li, Shen and Luo, Xiao and Qu, Xingda},
  journal={IEEE Transactions on Intelligent Vehicles},
  volume={8},
  number={2},
  pages={1025--1033},
  year={2022},
  publisher={IEEE}
}

@article{wang2024implicit,
  title={Implicit predictive behavior cloning for autonomous driving decision-making in urban traffic},
  author={Wang, Xudong and Wei, Chao and Tian, Hanqing and Wang, Weida and Hu, Jibin},
  journal={IEEE Transactions on Intelligent Vehicles},
  year={2024},
  publisher={IEEE}
}

@article{eraqi2022dynamic,
  title={Dynamic conditional imitation learning for autonomous driving},
  author={Eraqi, Hesham M and Moustafa, Mohamed N and Honer, Jens},
  journal={IEEE Transactions on Intelligent Transportation Systems},
  volume={23},
  number={12},
  pages={22988--23001},
  year={2022},
  publisher={IEEE}
}

@article{yuan2025model,
  title={Model-Free Deep Reinforcement Learning with Multiple Line-of-Sight Guidance Laws for Autonomous Underwater Vehicles Full-Attitude and Velocity Control},
  author={Yuan, Chengren and Shuai, Changgeng and Zhang, Zhanshuo and Ma, Jianguo and Fang, Yuan and Sun, YuChen},
  journal={Advanced Intelligent Systems},
  pages={2400991},
  year={2025},
  publisher={Wiley Online Library}
}

@inproceedings{radwan2021obstacles,
  title={Obstacles avoidance of self-driving vehicle using deep reinforcement learning},
  author={Radwan, Mahmoud Osama and Sedky, Ahmed Ahmed Hesham and Mahar, Khaled Mohammed},
  booktitle={2021 31st International Conference on Computer Theory and Applications (ICCTA)},
  pages={215--222},
  year={2021},
  organization={IEEE}
}

@inproceedings{mahmoudi2023reinforcement,
  title={Reinforcement Learning for Obstacle Avoidance Application in Unity Ml-Agents.},
  author={Mahmoudi, Reza and Ostreika, Armantas},
  booktitle={IVUS},
  pages={214--221},
  year={2023}
}

@inproceedings{pinto2021curriculum,
    author    = {Pinto, Felipe and Al-Stouhi, Sameer},
    title     = {Curriculum Reinforcement Learning for Autonomous Driving},
    booktitle   = {SAE World Congress Experience},
    year      = {2021},
    organization = {SAE International}
}

@inproceedings{lu2023imitation,
  title={Imitation is not enough: Robustifying imitation with reinforcement learning for challenging driving scenarios},
  author={Lu, Yiren and Fu, Justin and Tucker, George and Pan, Xinlei and Bronstein, Eli and Roelofs, Rebecca and Sapp, Benjamin and White, Brandyn and Faust, Aleksandra and Whiteson, Shimon and others},
  booktitle={2023 IEEE/RSJ International Conference on Intelligent Robots and Systems (IROS)},
  pages={7553--7560},
  year={2023},
  organization={IEEE}
}

@article{zhao2024survey,
  title={A survey on recent advancements in autonomous driving using deep reinforcement learning: Applications, challenges, and solutions},
  author={Zhao, Rui and Li, Yun and Fan, Yuze and Gao, Fei and Tsukada, Manabu and Gao, Zhenhai},
  journal={IEEE Transactions on Intelligent Transportation Systems},
  year={2024},
  publisher={IEEE}
}

@article{lanzaro2025evaluating,
  title={Evaluating driver-pedestrian interaction behavior in different environments via Markov-game-based inverse reinforcement learning},
  author={Lanzaro, Gabriel and Sayed, Tarek},
  journal={Expert Systems with Applications},
  volume={260},
  pages={125405},
  year={2025},
  publisher={Elsevier}
}

@article{huang2023conditional,
  title={Conditional predictive behavior planning with inverse reinforcement learning for human-like autonomous driving},
  author={Huang, Zhiyu and Liu, Haochen and Wu, Jingda and Lv, Chen},
  journal={IEEE Transactions on Intelligent Transportation Systems},
  volume={24},
  number={7},
  pages={7244--7258},
  year={2023},
  publisher={IEEE}
}

@inproceedings{wang2021decision,
  title={Decision making for autonomous driving via augmented adversarial inverse reinforcement learning},
  author={Wang, Pin and Liu, Dapeng and Chen, Jiayu and Li, Hanhan and Chan, Ching-Yao},
  booktitle={2021 IEEE International Conference on Robotics and Automation (ICRA)},
  pages={1036--1042},
  year={2021},
  organization={IEEE}
}

@article{schwonberg2023survey,
  title={Survey on unsupervised domain adaptation for semantic segmentation for visual perception in automated driving},
  author={Schwonberg, Manuel and Niemeijer, Joshua and Term{\"o}hlen, Jan-Aike and Schmidt, Nico M and Gottschalk, Hanno and Fingscheidt, Tim and others},
  journal={IEEE Access},
  volume={11},
  pages={54296--54336},
  year={2023},
  publisher={IEEE}
}

@article{chi2024dynamic,
  title={Dynamic obstacle avoidance model of autonomous driving with attention mechanism and temporal residual block},
  author={Chi, Xinrui and Guo, Zhanbin and Cheng, Fu},
  journal={Alexandria Engineering Journal},
  volume={105},
  pages={538--548},
  year={2024},
  publisher={Elsevier}
}

@article{ma2024adaptive,
  title={Adaptive Attention Module for Image Recognition Systems in Autonomous Driving},
  author={Ma, Xianghua and Hu, Kaitao and Sun, Xiangyu and Chen, Shining},
  journal={International Journal of Intelligent Systems},
  volume={2024},
  number={1},
  pages={3934270},
  year={2024},
  publisher={Wiley Online Library}
}

@article{lu2024epitester,
  title={Epitester: Testing autonomous vehicles with epigenetic algorithm and attention mechanism},
  author={Lu, Chengjie and Ali, Shaukat and Yue, Tao},
  journal={IEEE Transactions on Software Engineering},
  year={2024},
  publisher={IEEE}
}

@article{xi2023ema,
  title={EMA-GAN: A Generative Adversarial Network for Infrared and Visible Image Fusion with Multiscale Attention Network and Expectation Maximization Algorithm},
  author={Xi, Xiuliang and Jin, Xin and Jiang, Qian and Lin, Yu and Zhou, Wei and Guo, Lei},
  journal={Advanced Intelligent Systems},
  volume={5},
  number={11},
  pages={2300310},
  year={2023},
  publisher={Wiley Online Library}
}

@inproceedings{wang2024pedestrian,
  title={Pedestrian trajectory intention prediction in autonomous driving scenarios based on spatio-temporal attention mechanism},
  author={Wang, Yong and Wan, Weixiang and Zhang, Hanqing and Chen, Chen and Jia, Guancong},
  booktitle={2024 4th International Conference on Electronic Information Engineering and Computer Communication (EIECC)},
  pages={1519--1522},
  year={2024},
  organization={IEEE}
}

@article{peng2025diffusion,
  title={Diffusion models for intelligent transportation systems: A survey},
  author={Peng, Mingxing and Chen, Kehua and Guo, Xusen and Zhang, Qiming and Zhong, Hui and Zhu, Meixin and Yang, Hai},
  journal={IEEE Transactions on Intelligent Transportation Systems},
  year={2025},
  publisher={IEEE}
}

@article{wang2024trajectory,
  title={Trajectory grid diffusion for multimodal trajectory prediction in autonomous vehicles},
  author={Wang, Jincheng and Guo, Jiayu and Feng, Mingyue and Li, Chengjun and Xue, Xiangyang and Pu, Jian},
  journal={IEEE Transactions on Intelligent Vehicles},
  year={2024},
  publisher={IEEE}
}

@article{schulman2017proximal,
  title={Proximal policy optimization algorithms},
  author={Schulman, John and Wolski, Filip and Dhariwal, Prafulla and Radford, Alec and Klimov, Oleg},
  journal={arXiv preprint arXiv:1707.06347},
  year={2017}
}

@article{ho2016generative,
  title={Generative adversarial imitation learning},
  author={Ho, Jonathan and Ermon, Stefano},
  journal={Advances in neural information processing systems},
  volume={29},
  year={2016}
}

@inproceedings{wang2023diffusion,
  author = {Wang, Zhendong and Janner, Michael and Du, Yilun and Finn, Chelsea and Levine, Sergey},
  title = {Diffusion Policies as an Expressive Policy Class for Offline Reinforcement Learning},
  booktitle = {International Conference on Learning Representations},
  year = {2023}
}

@article{pearce2023plannable_journal,
    author    = {Pearce, Tim and Canonaco, Alessandro and Zhu, He and Havoutis, Ioannis and Pugeault, Nicolas},
    title     = {Imitating Human Behaviour with Diffusion Models},
    journal   = {IEEE Robotics and Automation Letters},
    year      = {2023},
    volume    = {8},
    number    = {6},
    pages     = {3607-3614}
}

@inproceedings{chi2023diffusion,
  title={Diffusion-based generation, optimization, and planning in 3d scenes},
  author={Chi, Cheng and Little, Vicki and Zhang, Siyuan and Wang, Zhaoyang and Wang, Shu and Chang, Angel and Sung, Manolis},
  booktitle={Conference on Robot Learning},
  pages={2209--2220},
  year={2023}
}

@inproceedings{janner2022planning,
  title={Planning with Diffusion for Flexible Behavior Synthesis},
  author={Janner, Michael and Du, Yilun and Tenenbaum, Joshua B and Levine, Sergey},
  booktitle={Proceedings of the 39th International Conference on Machine Learning (ICML)},
  pages={9483--9496},
  year={2022},
  series={Proceedings of Machine Learning Research}
}

@article{ho2020denoising,
  title={Denoising diffusion probabilistic models},
  author={Ho, Jonathan and Jain, Ajay and Abbeel, Pieter},
  journal={Advances in neural information processing systems},
  volume={33},
  pages={6840--6851},
  year={2020}
}
\vspace{-41pt}  

\begin{IEEEbiography}[{\includegraphics[width=1in, height=1.25in, clip, keepaspectratio]{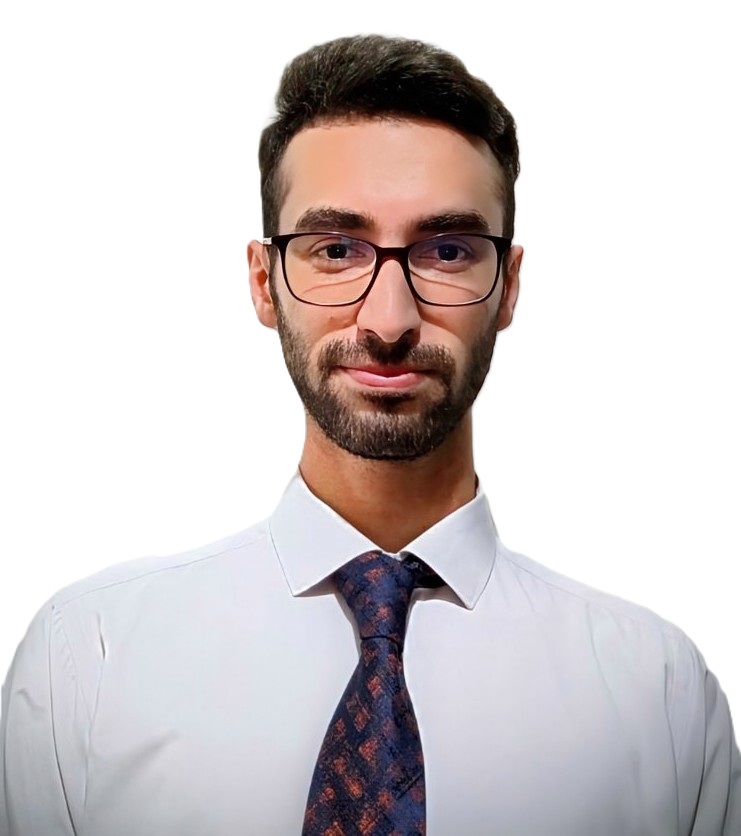}}]{Seyed Ahmad Hosseini Miangoleh}
is currently pursuing the B.S. degree in electrical engineering (control) at Amirkabir University of Technology, Tehran, Iran. His research interests include control systems, robotics, machine learning, deep learning, reinforcement learning, and computer vision.
\end{IEEEbiography}
\vspace{-52pt}  

\begin{IEEEbiography}[{\includegraphics[width=1in, height=1.25in, clip, keepaspectratio]{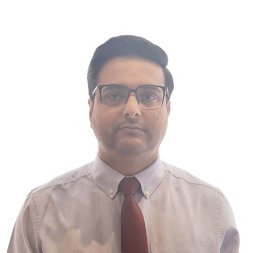}}]{Amin Jalal Aghdasian}
received the B.S. degree in electrical engineering from Tabriz University, Tabriz, Iran, in 2018, and the M.S. degree in mechatronics engineering from Amirkabir University of Technology, Tehran, Iran, in 2023. His research interests include intelligent automotive systems, robust control, machine learning, neural networks, robotics, and reinforcement learning.
\end{IEEEbiography}
\vspace{-53pt}  

\begin{IEEEbiography}[{\includegraphics[width=1in, height=1.25in, clip, keepaspectratio]{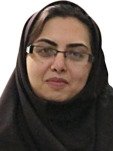}}]{Farzaneh Abdollahi}
(Senior Member, IEEE) received the B.Sc. degree in electrical engineering from the Isfahan University of Technology, Isfahan, Iran, in 1999, the M.Sc. degree in electrical engineering from the Amirkabir University of Technology, Tehran, Iran, in 2003, and the Ph.D. degree in electrical engineering from Concordia University, Montreal, QC, Canada, in 2008. She is currently an Associate Professor with the Amirkabir University of Technology and an adjunct Professor with Carleton University. Her research interests include intelligent control, robotics, control of nonlinear systems, control of multiagent networks, and robust and switching control.
\end{IEEEbiography}

\end{document}